\documentclass[10pt,letterpaper]{article}

\usepackage{cogsci}

  \cogscifinalcopy 
 \usepackage[table,xcdraw]{xcolor}
\usepackage{graphicx}
\usepackage{amsmath,amssymb}
\usepackage{algorithm}
\usepackage{algorithmic}
\usepackage{pslatex}
\usepackage{apacite}
\usepackage{soul}
\usepackage{subcaption}
\usepackage{float} 
                   
\usepackage[round]{natbib}
\newcommand{\mfunc}{\mathcal{L}}

\newcommand{\C}{\mathcal{C}}

\usepackage{lipsum}

\title{Pragmatic Reasoning in Structured Signaling Games}
\author{{\large \bf Emil Carlsson (caremil@chalmers.se)} \\
  \AND {\large \bf Devdatt Dubhashi (dubhashi@chalmers.se)} \\
  Department of Computer Science, Chalmers University of Technology\\
  Gothenburg, 412 96 Sweden}

\date{December 2021}

\begin{document}

\maketitle

\begin{abstract}
In this work we introduce a structured signaling game, an extension of the classical signaling game with a similarity structure between meanings in the context, along with a variant of the Rational Speech Act (RSA) framework which we call structured-RSA (sRSA) for pragmatic reasoning in structured domains. We explore the behavior of the sRSA in the domain of color and show that pragmatic agents  using sRSA on top of semantic representations, derived from the World Color Survey, attain efficiency very close to the information theoretic limit after only 1 or 2 levels of recursion. We also explore the interaction between pragmatic reasoning and learning in multi-agent reinforcement learning framework. Our results illustrate that artificial agents using sRSA develop communication closer to the information theoretic frontier compared to agents using RSA and just reinforcement learning. We also find that the ambiguity of the semantic representation increases as the pragmatic agents are allowed to perform deeper reasoning about each other during learning. 


\end{abstract}

\section{Introduction}
The Rational Speech Act (RSA) framework \citep{Frank12, Goodman2016} has emerged as a leading probabilistic model of pragmatic communication formalizing the Gricean view on pragmatics \citep{Grice1975}. In RSA models, each agent reasons about the other agent's belief, in a game-theoretic fashion, in order to infer the context dependent meaning of an utterance. Models of this type have been used to make accurate predictions about human behavior over a wide range of different and complex tasks \citep{Goodman2016}.



It was recently shown by \citet{Peloquin2020} that efficient language use and structure emerge as pragmatic agents interact with each other in a signaling game. In their framework the efficiency was measured as the expected cross-entropy between the speaker and listener distributions.

However, in certain settings, the meaning space may have special structure which needs to be exploited to develop efficient communication. A good example is the domain of colors where it is possible to quantify the similarity between different colors. Hence, in a context where agents are talking about different colors an error might be quantified differently depending on whether the listener confused the color the speaker was referring to with a very similar color or with a completely different color. This is something that is not captured by a purely entropy-based efficiency measure. 

Here we take a new approach to the basic question addressed in \citet{Peloquin2020} about how efficient communication arises via the interaction of pragmatic agents.
First, to take structure into account, we introduce a notion of a \emph{structured signaling game}, an extension of the standard signaling game, commonly used in work regarding pragmatic reasoning. 
For this type of signaling game we introduce an extension of the standard RSA which we call \emph{structured-RSA} (sRSA) where an agent accounts for the structure in the meaning space during the reasoning process. We explore the differences between RSA and sRSA in the color domain, a domain commonly used in cognitive science to explore various linguistic phenomena \citep{Regier2015, Gibson2017}.
Second, we quantify the efficiency of the resulting communication schemes using the information theoretic notions of efficiency from \citet{Zaslavsky2018a} and the well-formedness measure from \citet{Regier2007}.

We first investigate the use of human representations such as the color naming systems found in the World Color Survey \citep{Cook} as a basis for reasoning by pragmatic agents. We show that efficiency of communication increases much more when agents reason using sRSA compared to agents using RSA and base policies. The most striking result is that sRSA agents initialized with human representations only need a recursion depth of $1$ or $2$ in order to come very close to the optimal frontier. 

\begin{figure*}
    \centering
    \vspace{-1.5cm}
    \includegraphics[width=\textwidth, trim={3cm 4cm 3cm 2.5cm},clip]{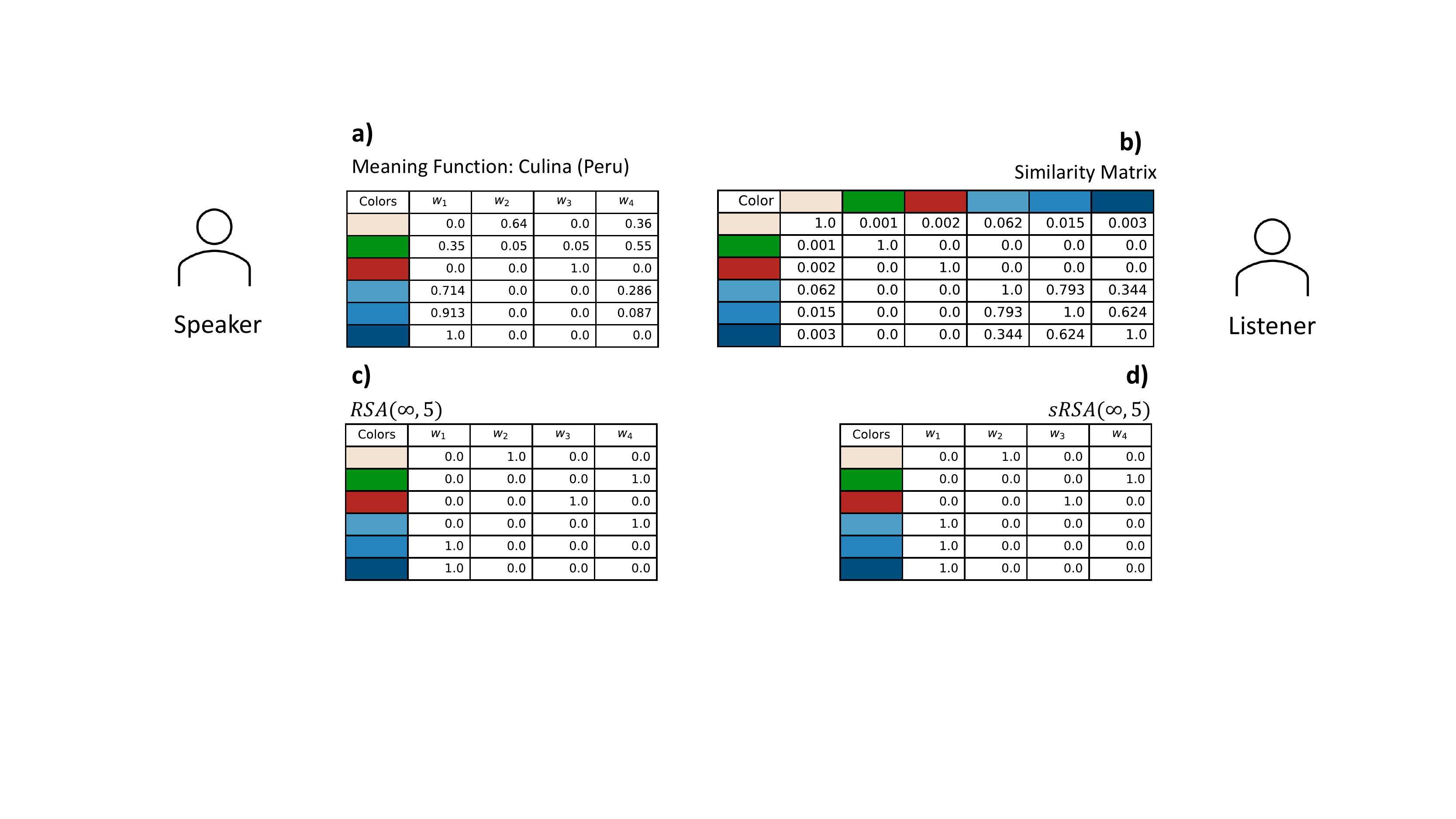}
    \caption{An example of a structured signaling game in the color domain.} 
    \label{fig:structured_signaling_game}
\end{figure*}

Next, we consider computational learning agents interacting with each other in a multi-agent reinforcement learning framework similar to those considered in \citet{Kageback2020, Chaabouni21, Carlsson21,  Ohmer22}. Our results in this learning framework suggest that pragmatic agents equipped with sRSA learn more efficient color naming systems compared to agents using RSA or pure reinforcement learning. We also find that ambiguity arises to a greater extent in the semantic representation as the computational agents are allowed to perform deeper reasoning about each other. Even though the ambiguity increases, the computational agents using sRSA still develop efficient and accurate communication. Compared to previous  works \citep{Monroe17, Kageback2020, Chaabouni21, hu_scalable_2021}, which only account for the structure of the color space in the non-contextual meaning function. Our approach extends this and explicitly accounts for structure in the RSA recursion. 

The work of \citet{Zaslavsky2020rdrsa} is also related to our work. They use the fact that the softmax operator maximizes a trade-off between utility and entropy \citep{Fudenberg1998} to argue that the RSA recursion can be viewed as an alternating maximization of a least-effort objective. They ground the recursion in Rate-Distortion theory and derive a new update of the sender based on the mutual information between meaning and utterance. In contrast to their work, our sRSA is based on the standard RSA recursion, with the difference that our utility function leverages the pair-wise similarity, or distortion, between meanings in the context.

\section{Structured Signaling Games and sRSA}
In our signaling game, two agents, one sender and one listener, observe a context of $n$ meanings $\C = \{m_i\}$ where each $m_i$ lies in some meaning space $\mathcal{M}$. The goal of the sender is to describe one of the meanings to the listener. In the standard setup of a signaling game, the agents share a semantic representation, or meaning function, $\mfunc(m, w)$, which describes how well the utterance $w$ describes the object $m$. In our structured version we also assume that the agents share a similarity matrix $Z$ where element $Z_{ij}$ describes how similar meanings $m_i$ and $m_j$ are. We assume $Z_{ij} \in [0, 1]$ with $Z_{ii}=1$. An example of a structured signaling game in the domain of colors is presented in Figure \ref{fig:structured_signaling_game}.

\begin{figure*}[t]
    \centering
    \vspace{-2cm}
\begin{tabular}{cc}
    \subfloat[Information-theoretic trade-off between complexity and accuracy after one recursion.]{\includegraphics[width=0.5\textwidth]{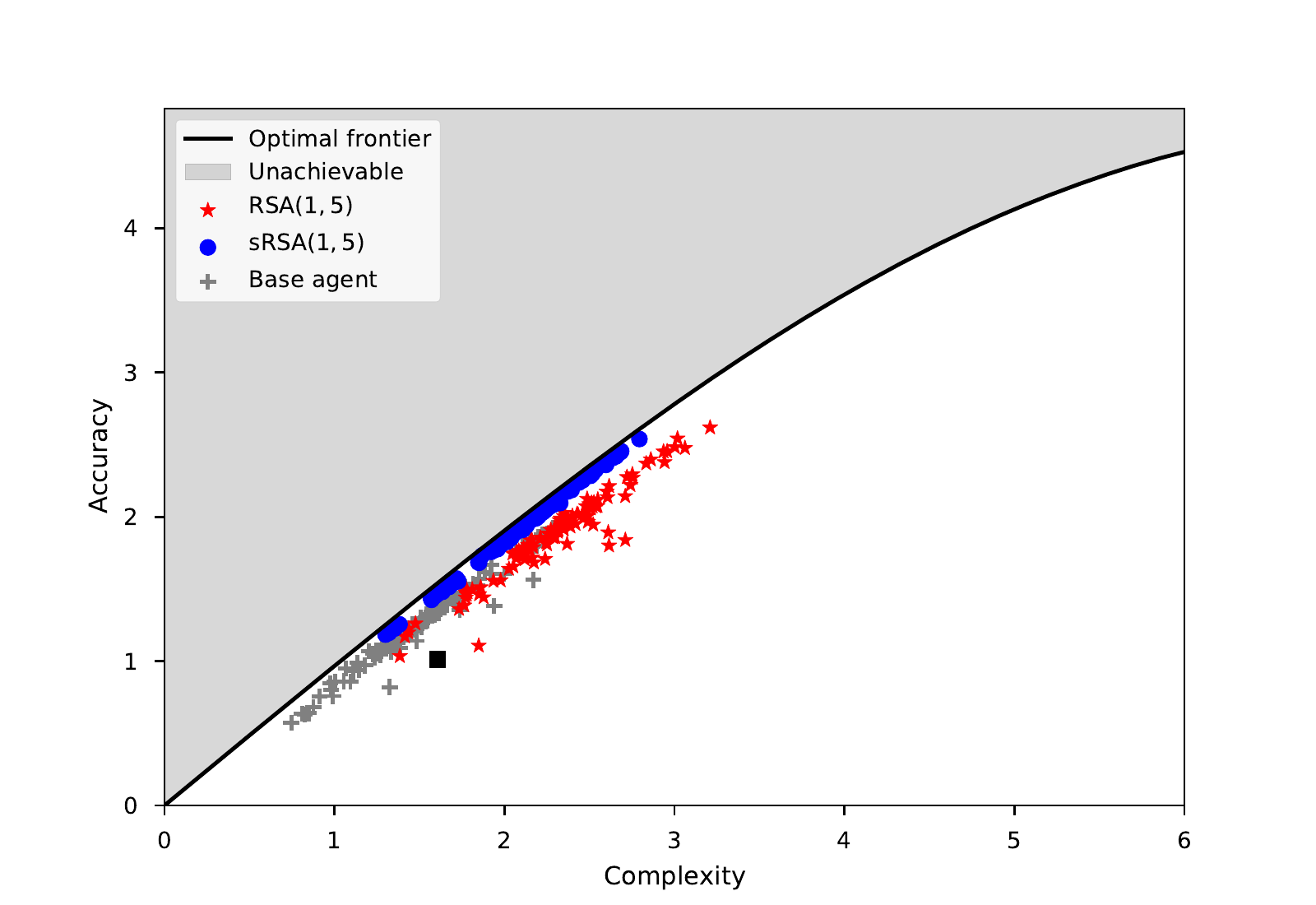}\label{fig:human_IB}} 
    &
    \subfloat[Well-formedness of the agents after one recursion.]{\includegraphics[width=0.5\textwidth]{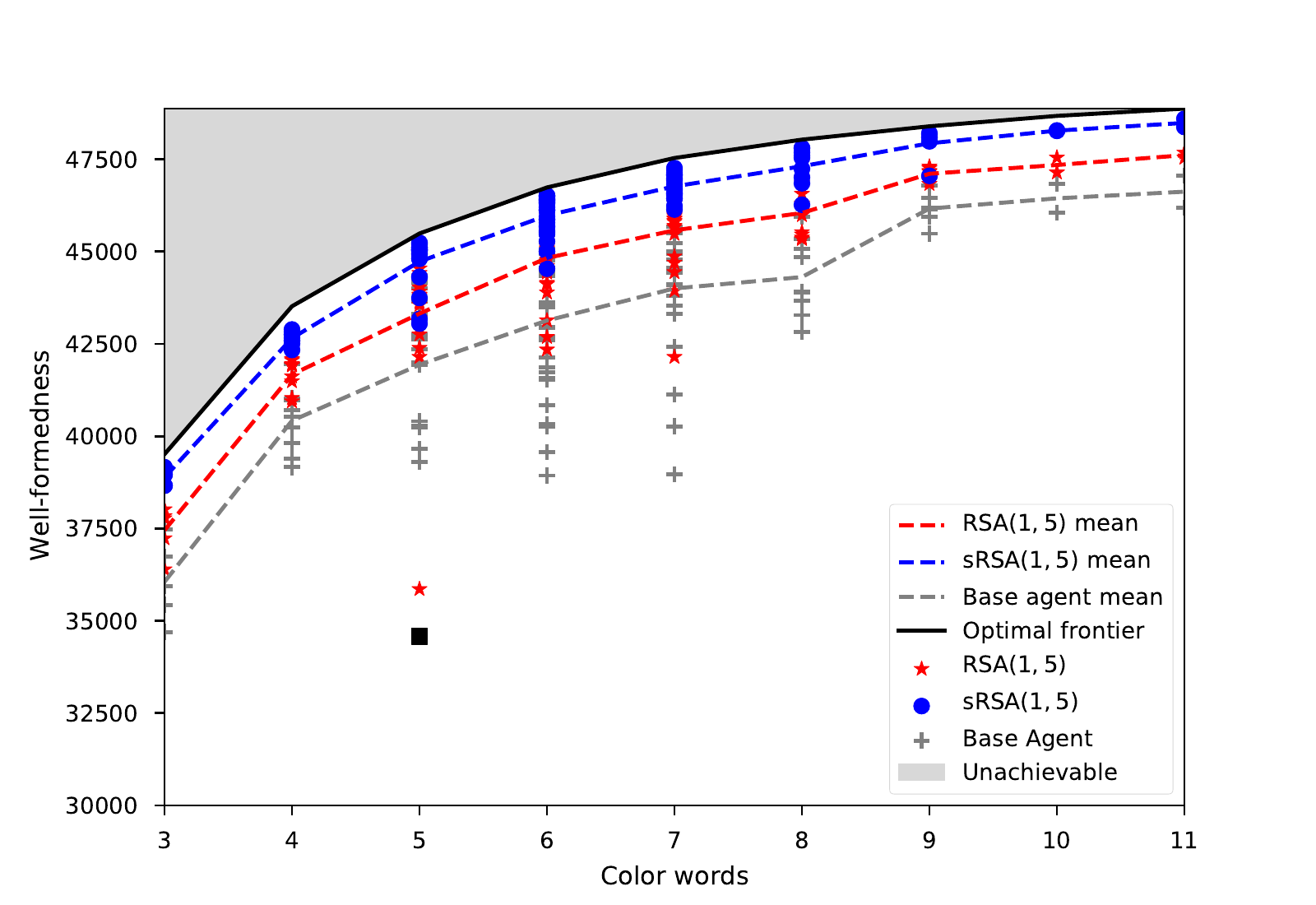}\label{fig:well_depth1}}
    \\
    \subfloat[Information-theoretic trade-off between complexity and accuracy in the limit.]{\includegraphics[width=0.5\textwidth]{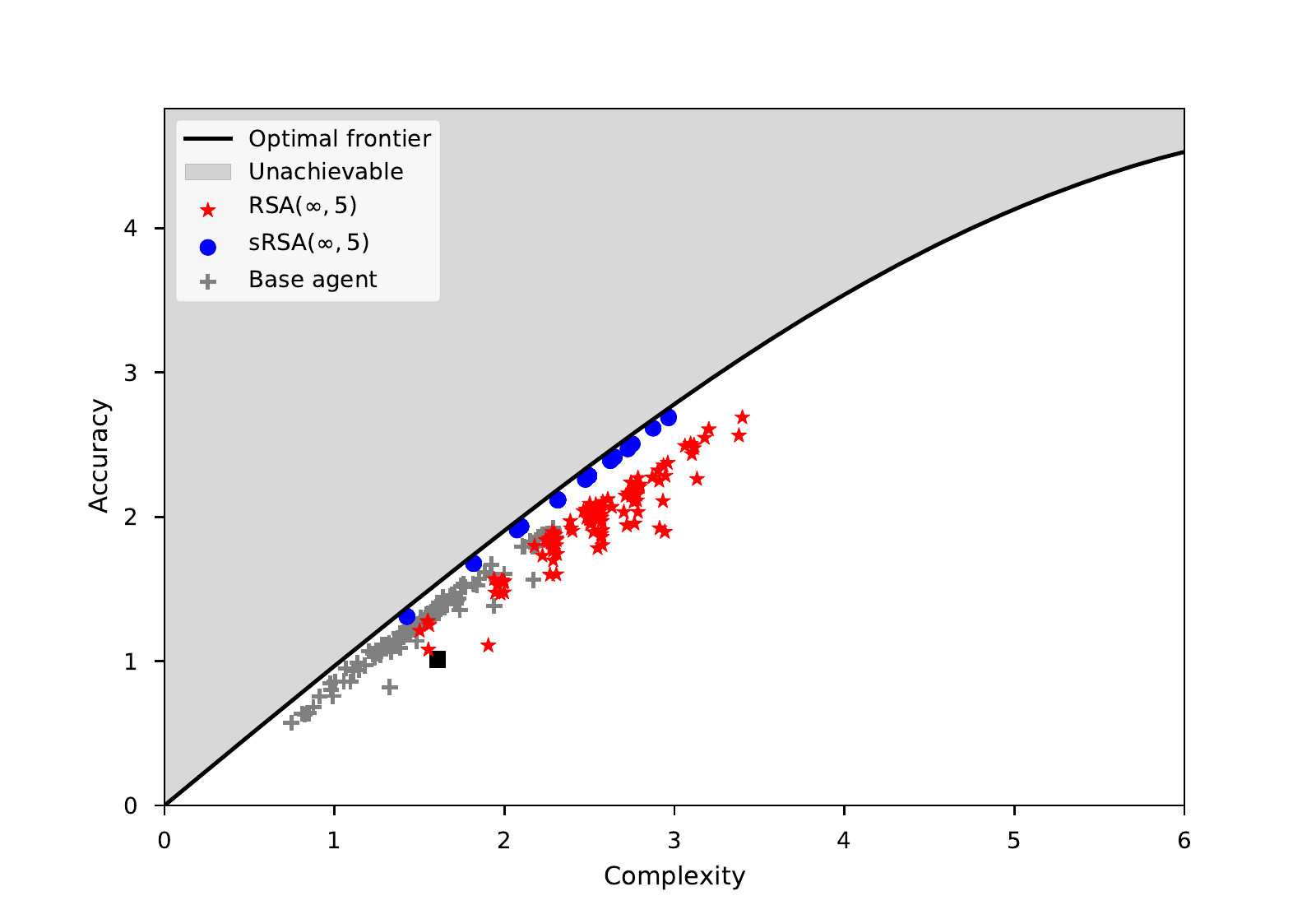}\label{fig:human_IB_inf}}
    &
    \subfloat[Well-formedness as the recursions goes to infinity.]{\includegraphics[width=0.5\textwidth]{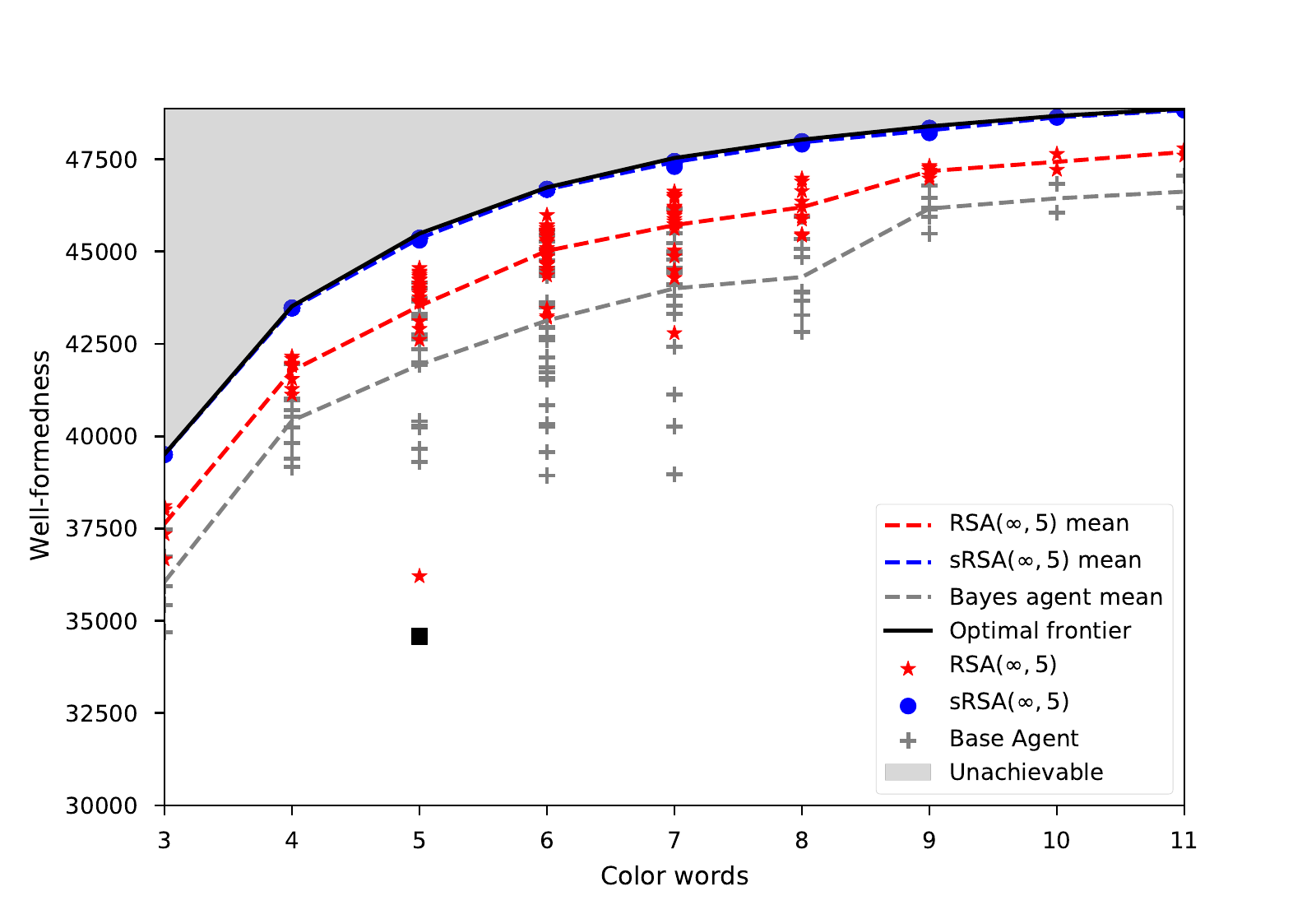}\label{fig:human_well_inf}}
\end{tabular}
    \caption{Results for applying pragmatic reasoning on-top of the color naming data in WCS. Here depth of recursion indicates the depth of the final sender in the recursion. We use $\alpha=5$ in the recursions. The black square  indicates the position of the base agent of the language Karajá.}
    \label{fig:human_exp}
\end{figure*}

\subsection{Similarity-Sensitive Utility and sRSA}
Following \citet{Degen20}, we consider agents equipped with a continuous meaning function, or semantic representation, 
$\mfunc(m,w) \in [0, 1]$
which describes how well a meaning $m$ can be mapped to an utterance $w$.
On top of the meaning function, our agents use the RSA in order to reason about each other's behavior given the context $C$. Given a literal listener proportional to the meaning function, $L_0(m|w) \propto \mfunc(m,w) $, the following recursion is applied in the RSA
\begin{align}
    S_t(w|m, C) &\propto e^{\alpha U_{t}(m, w, C)} \\
    L_t(m|w, C) &\propto S_t(w|m, C)p(m|C)
\end{align}
where $U_{t}(w, m, C)$ is the expected utility, of conveying message $w$ given the meaning $m$ in the context $C$, and $p(m|C)$ is the prior probability of $m$ given $C$. In RSA the utility of the sender is usually based on reducing the epistemic uncertainty the listener carries about the true meaning, and is taken to be the negative surprisal of the listener $U_{t}(w, m, C) = \log L_{t-1}(m|w, C)$. We will denote an agent using RSA at a recursion depth of $t$ with parameter $\alpha$ as \emph{RSA$(t, \alpha)$}.

\subsubsection{Similarity-Sensitive Surprisal}
\citet{Lein21} recently introduced extensions of entropy and other information theoretic concepts in the context of structured domains, where one has a matrix of similarities $Z$. Inspired by this, we define the \emph{similarity-sensitive surprisal} of a listener, $L$, as \begin{align}\label{eq:sim-sup}
I^Z(m, w, C) = - \log \sum_{m'} Z_{m m'}L(m'|w, C).
\end{align}
Here $Z(m,m')$ is the similarity between the two meanings $m$ and $m'$. This measure captures the desirable property that a listener shouldn't be as surprised if a speaker used the same word for two similar colors compared to if the speaker used the same word for two very different colors.

Defining the utility as $U(m, w) = - I^{Z}(m, w)$ we arrive at structured version of RSA (sRSA) with similarity-sensitive sender. Note that this utility yields a sender proportional to the power $\alpha$ of the expected similarity \begin{align}\label{eq:sim-sender}
    S_t(w|m, \mathcal{C}) \propto (\sum_{m' \in \mathcal{C}}Z_{m m'}L_{t-1}(m'|w))^\alpha.
\end{align}
In next section~\ref{sec:srsa_v_rsa} and in Figure \ref{fig:structured_signaling_game} we give a simple example in the color domain to illustrate the difference between RSA and sRSA.

In the special case where $Z$ is the identity matrix, i.e. where meanings in the context share no similarity, (\ref{eq:sim-sup}) reduces to the standard surprisal and the sender in (\ref{eq:sim-sender}) reduces to the standard RSA sender. We will denote an agent using sRSA at a recursion depth of $t$ with parameter $\alpha$ as \emph{sRSA$(t, \alpha)$}.

In general, given a distortion measure on the meaning space  
$d: \mathcal{M} \times \mathcal{M} \rightarrow \mathbb{R}^+$,
we can construct a natural similarity measure as $Z_{mm'} := e^{-\beta d(m, m')}, \, \beta > 0$.

\section{Color Domain: Efficiency and Well-formedness}
We will use colors as our testbed for pragmatic reasoning in structured signaling games. The seminal work of  \citet{Zaslavsky2018a} showed that color naming systems in the World Color Survey (WCS) \citep{Cook} optimize an information-theoretic trade-off between complexity and accuracy of the meaning function. Following \citet{Zaslavsky2018a} we will take the complexity of a color naming system as the mutual information between word and meaning
\begin{align*}
   \text{Complexity} =  I(M;W)
\end{align*}
and the accuracy as \begin{align*}
    \text{Accuracy} = I(W;U).
\end{align*}
As in \citet{Zaslavsky2018a} we assume a meaning $m$ to be a distribution over color chips proportional to a isotropic Gaussian, $m(u) \propto e^{-\frac{1}{64}||x_m - x_u||^2}$ where $x_m$ is the CIELAB vector corresponding to color chip $m$.


\citet{Regier2007} showed also that human color naming reflects optimal partitions of the color space w.r.t. to a measure of \emph{well-formedness}. The well-formedness criterion was based on the following measure of perceptual similarity between colors 
\begin{align}\label{eq:sim}
    \text{sim}(m, m') = e^{-0.001||x_m - x_{m'}||^2}
\end{align}
This similarity measure will be used in our sRSA model in the downstream analysis.
\subsubsection{sRSA vs RSA}\label{sec:srsa_v_rsa}
Figure~\ref{fig:structured_signaling_game} gives a simple example of a structured signaling game where the context consists of $6$ different colors. The meaning function mapping color to word is based on the naming data found in the World Color Survey for the language Culina is shown in Figure 1a.  The similarity matrix, which describes how similar two colors are w.r.t. the similarity measure defined in (\ref{eq:sim}), is shown in Figure 1b. We use $RSA(t, \alpha)$ to denote the result of applying depth $t$ RSA and $RSA(\infty, \alpha)$ to denote the limit as $t \rightarrow \infty$, and similarly for sRSA. Figure 1c and Figure 1d show the limit points for RSA and sRSA (with $\alpha = 5$). Since RSA minimizes only the surprisal of the listener and does not account for the similarity structure we observe that the lighter blue color and green color are mapped to the same word. Unlike RSA, the sRSA takes the similarity matrix into account and converges to a solution where the first $3$ colors can be uniquely determined, while the last $3$, all variants of blue, are mapped to the same word. 

\subsection{Human Representations}
\begin{figure}[h]
    \centering
\begin{tabular}{cc}
    \subfloat[Efficiency trade-off]{\includegraphics[width=0.25\textwidth]{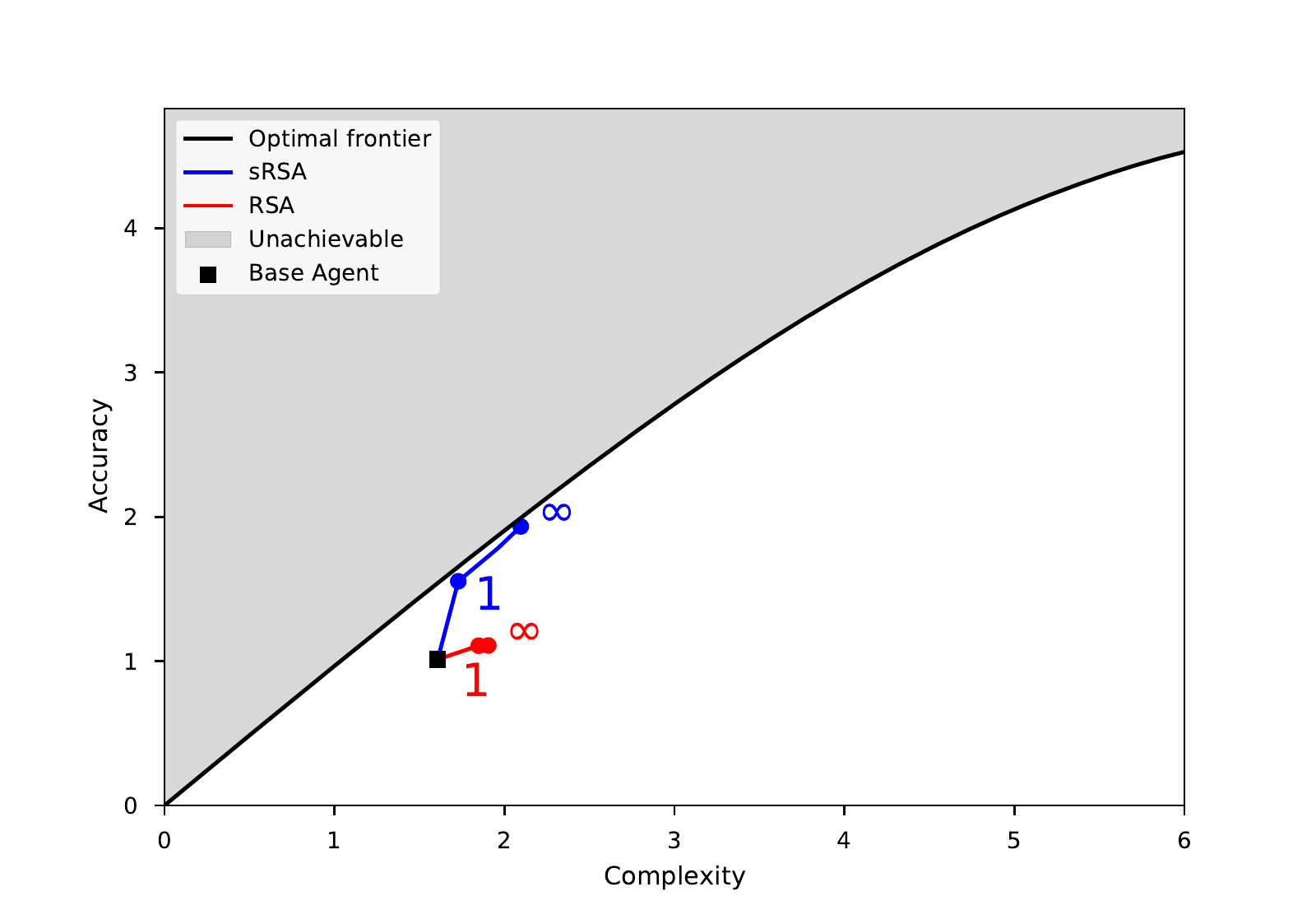}\label{fig:Karaja_efficiency}}
    & \hspace{-0.5cm}
    \subfloat[Well-formedness]{\includegraphics[width=0.25\textwidth]{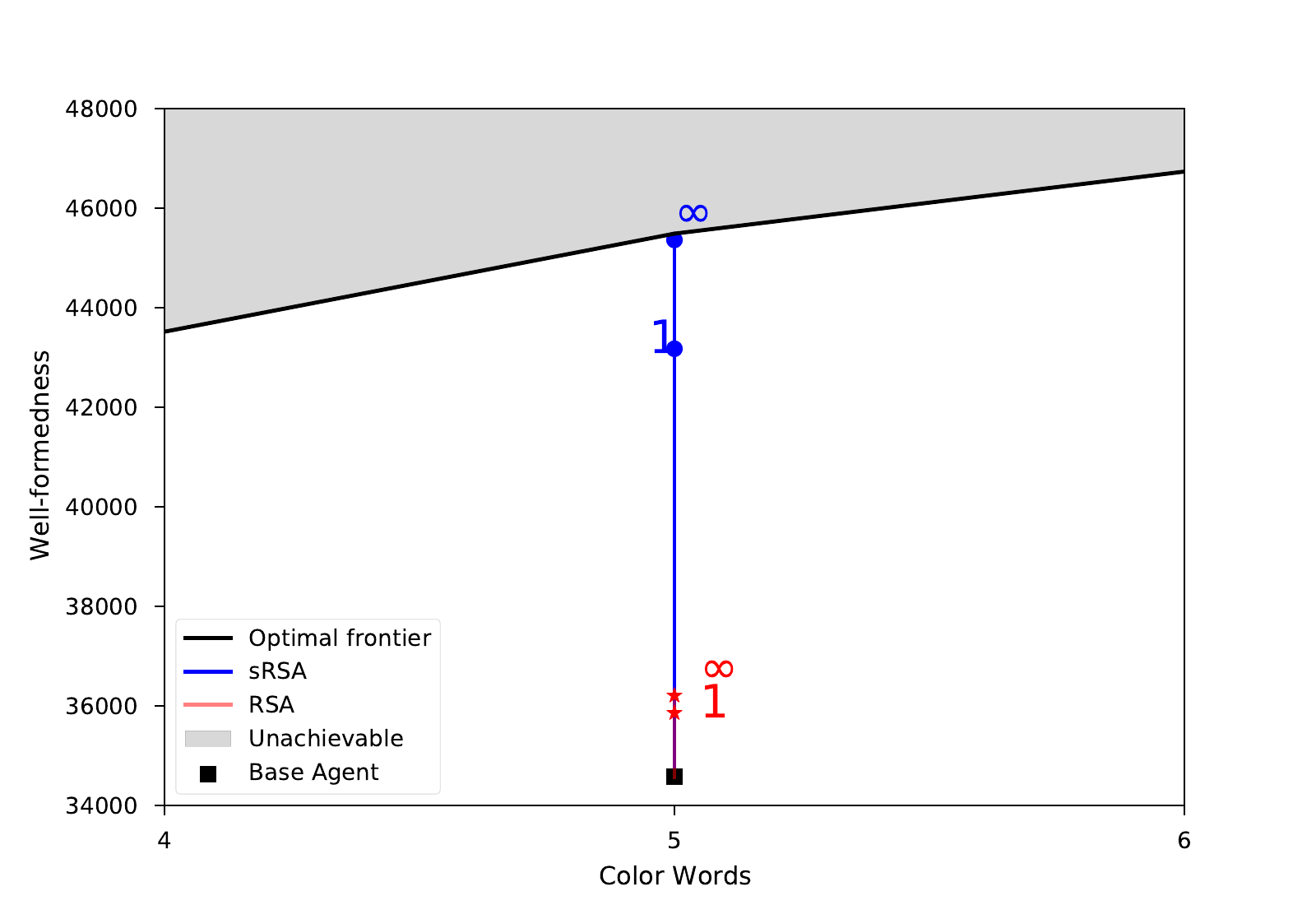}}
\end{tabular}
    \caption{Trajectories of RSA and sRSA for Karajá.}
    \label{fig:Karaja_plot}
\end{figure}
The WCS data consist of naming data from $110$ languages, with an average of $25$ speakers for each language. Since the WCS data contain data from speakers, we believe it is more appropriate to consider a slightly different version of the RSA recursion, where the agents start reasoning from a literal sender proportional to the naming data from WCS\footnote{As in \citet{Regier2015}, we only consider major color terms. We say that a color term is major if it is the mode category for at least $10$ chips in the Munsell Chart.}. For a language $l$ in the WCS study and corresponding naming data $D^l(w, m)$ we consider the following recursion \begin{align*}
    S_0^{l}(w, m, C) &\propto D^l(w, m) \\
    L_t^{l}(m|w, C) &\propto S_{t-1}^{l}(w|m, C)p(m|C)\\
    S_t^{l}(w|m) &\propto e^{U_{t}(w, m, C)}.
\end{align*}

We consider a structured signaling game with the context, $\mathcal{C}$, being the entire Munsell chart. Hence, a sender is given a certain color chip from the Munsell chart and should describe this to the listener, which then produces a distribution over the color chips in the chart. The context we consider here is much larger compared to the ones considered in, for example, \citet{Monroe17}. The reason is that we are interested in larger contexts where the number of meanings is much larger than the number of utterances and exact communication is impossible. We will consider a uniform need distribution over the chart and leave it for future work to study skewed priors like the one used in \citet{Zaslavsky2018a}. As a baseline we will consider the base agents from the recursion, i.e. a sender proportional to the naming data and the corresponding Bayesian listener. The information-theoretic frontier is computed using the Blahut-Arimoto algorithm with the annealing scheme outlined in \citet{Zaslavsky2018a} and a uniform prior. The well-formedness frontier is computed using the Correlation Clustering approach described in \citet{Kageback2020}.


In Figure \ref{fig:human_IB} we compare the efficiency of the base agents to the efficiency of the pragmatic agents after performing one recursion in the respective reasoning model. We observe that \emph{pragmatic reasoning leads to more complex and accurate behavior for both RSA and sRSA} compared to the base agents. However, we also observe that the RSA agents have not moved closer to the optimal frontier while the sRSA agents are very close to the frontier \emph{after only one recursion}. Interestingly, when the recursions are allowed to go the limit, Figure \ref{fig:human_IB_inf}, the RSA agents seem to move away from the optimal frontier while the sRSA converges to naming distributions very close to the optimal frontier.

\begin{figure}[h]
\centering
\begin{tabular}{cc}
     \subfloat[Base agent]{\includegraphics[width=0.24\textwidth,
     trim={0cm 3cm 0cm 3cm},clip]{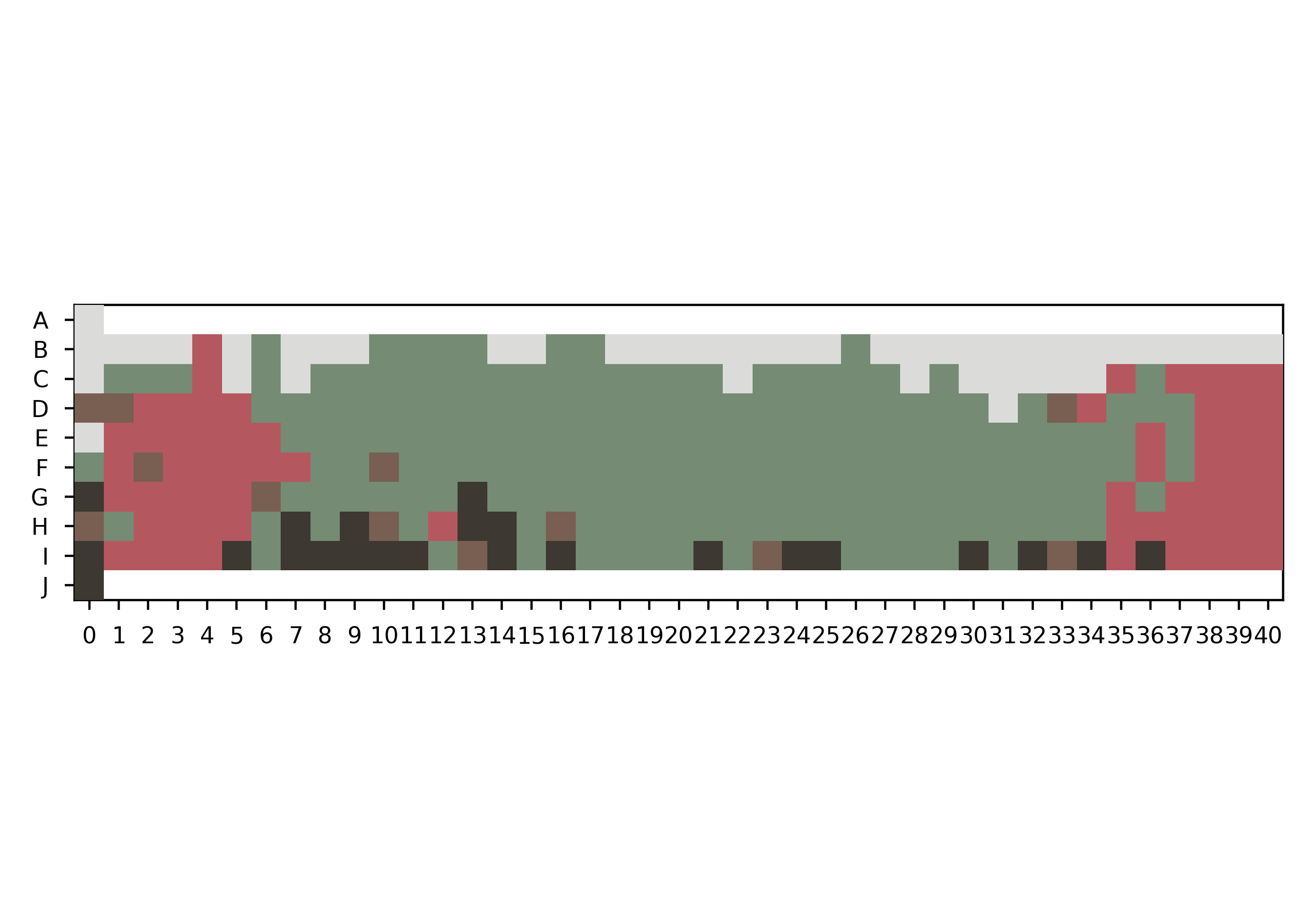}} 
     & \subfloat[CC agent]{\includegraphics[width=0.24\textwidth, 
     trim={0cm 3cm 0cm 3cm},clip]{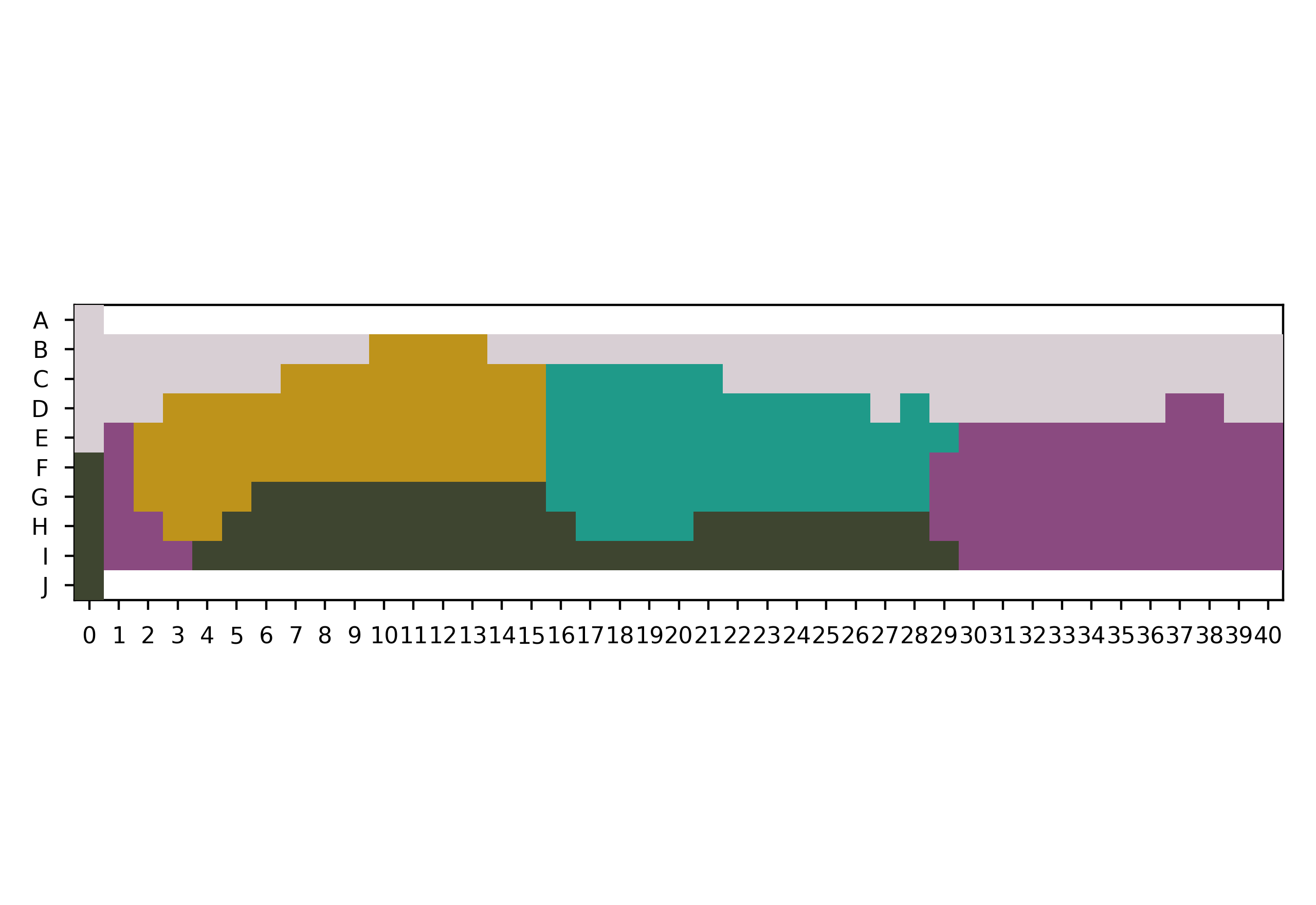}\label{fig:Karaja_ccc}}
     \\
     \subfloat[RSA$(1, 5)$]{\includegraphics[width=0.24\textwidth, trim={0cm 3cm 0cm 3cm},clip]{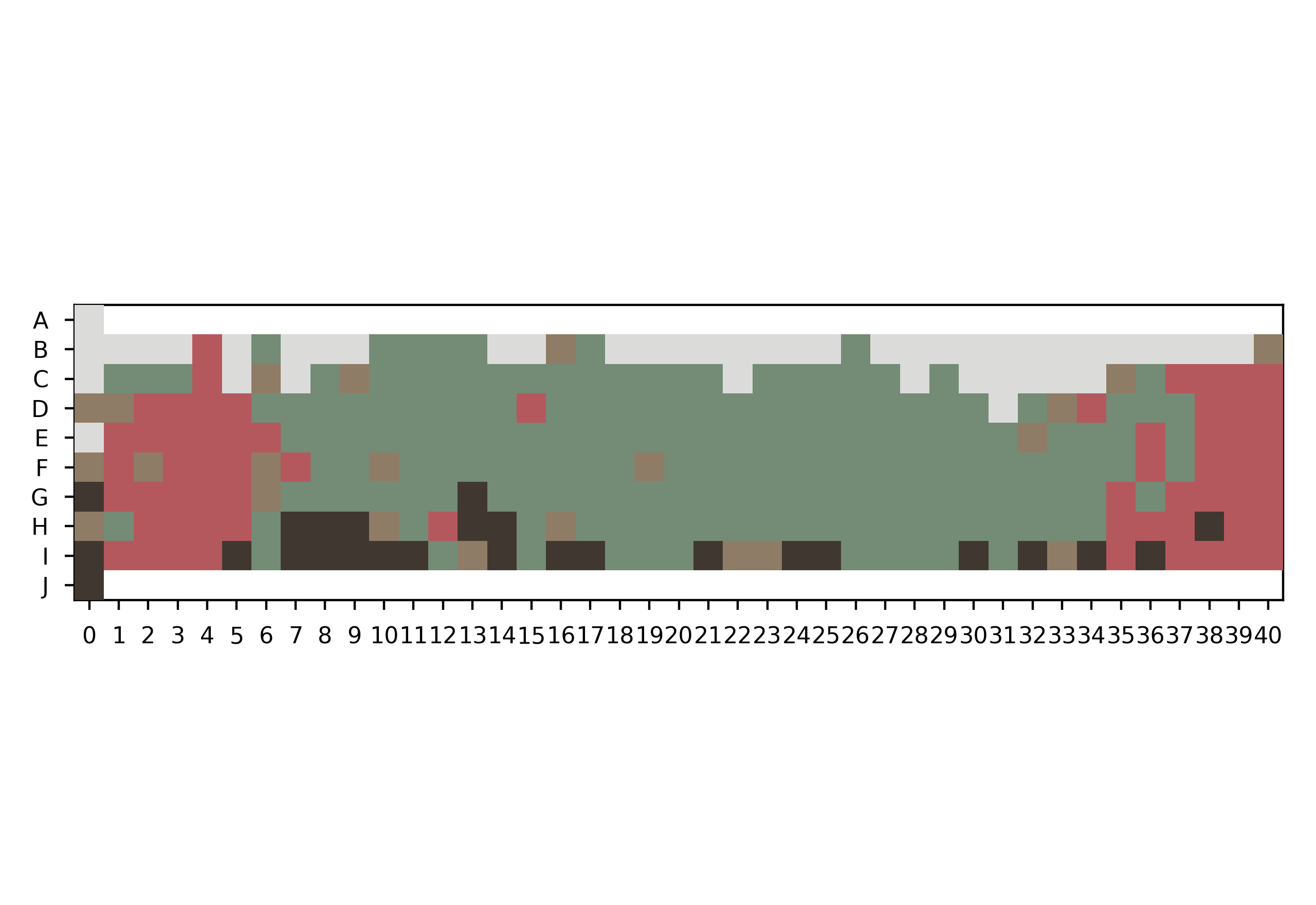}\label{fig:Karaja_rsa1}}
     &
     \subfloat[RSA$(\infty, 5)$]{\includegraphics[width=0.24\textwidth, trim={0cm 3cm 0cm 3cm},clip]{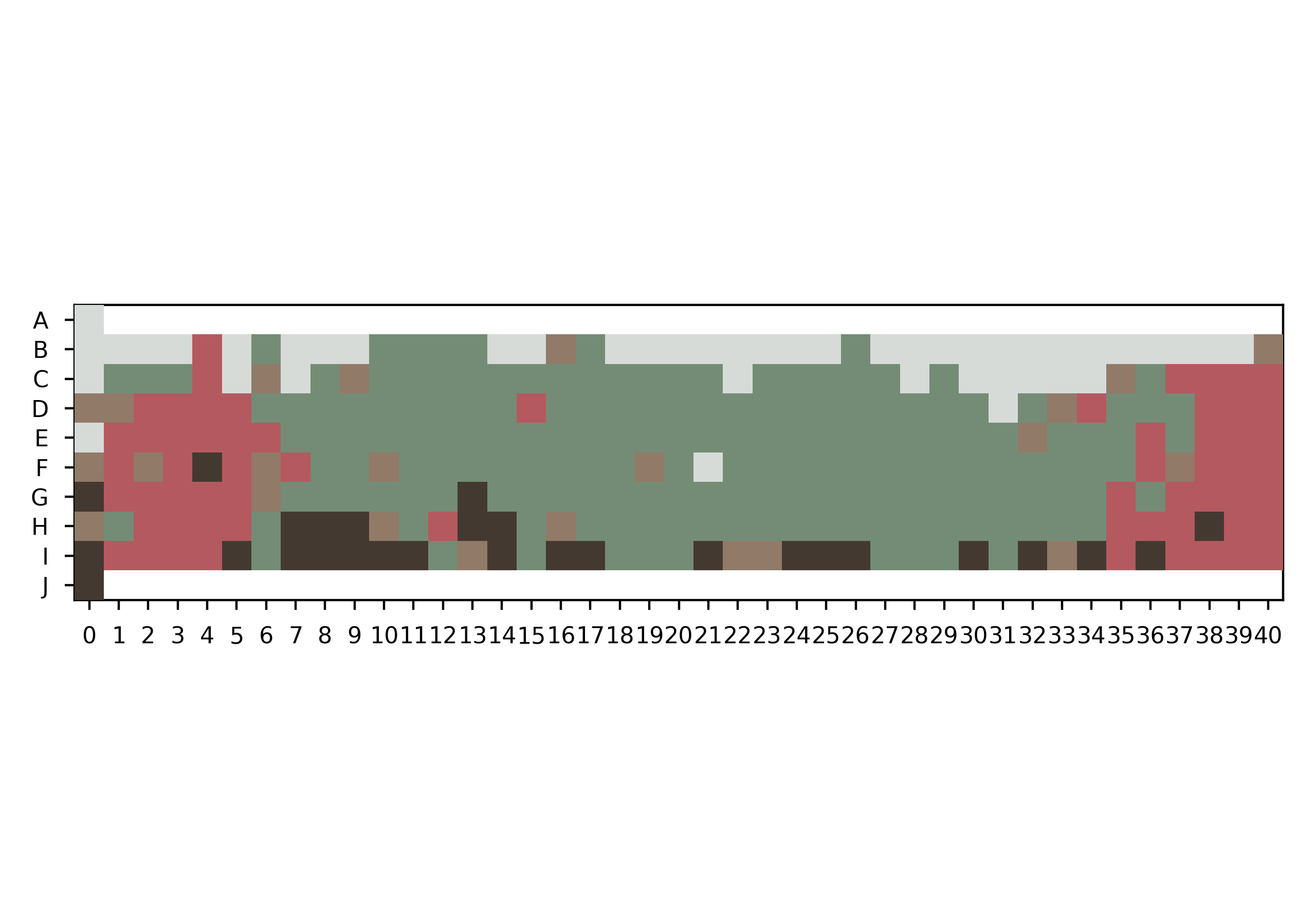}\label{fig:Karaja_rsa101}}
     \\
    \subfloat[sRSA$(1, 5)$]{\includegraphics[width=0.24\textwidth, trim={0cm 3cm 0cm 3cm},clip]{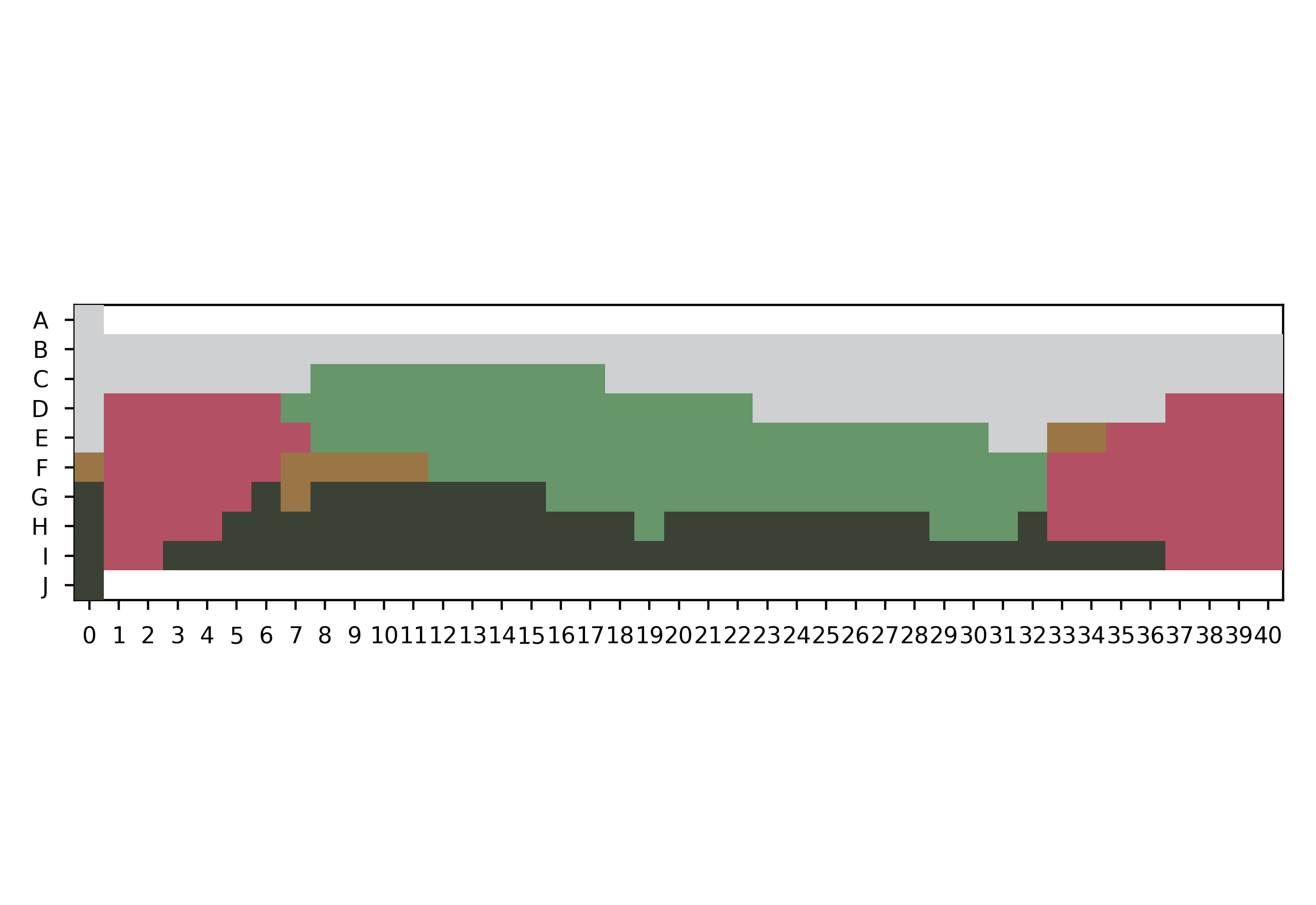}\label{fig:Karaja_rkk1}}
     &
     \subfloat[sRSA$(\infty, 5)$]{\includegraphics[width=0.24\textwidth, trim={0cm 3cm 0cm 3cm},clip]{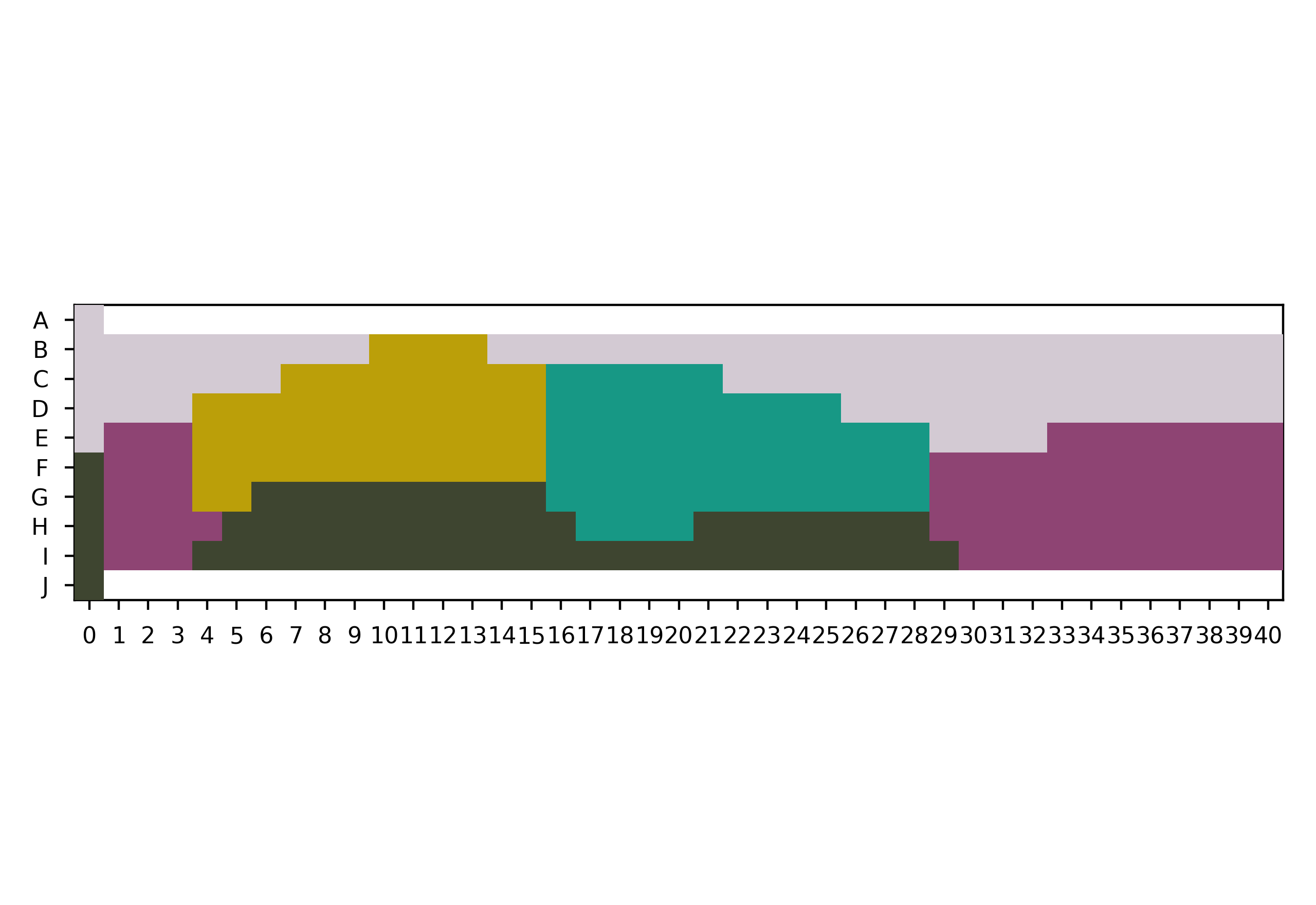}\label{fig:Karaja_rkk101}}
\end{tabular}
    \caption{Karajá, Brazil. The sRSA model refine and smooth the colormap in only one recursion. In the limit, we observe that the sRSA approaches the true optimal agent w.r.t. well-formedness (CC Agent). Each color term is colored with the average color mapped to the term. }
    \label{fig:Karaja}
\end{figure}

Further, Figure \ref{fig:well_depth1} illustrates the well-formedness of the agents after one recursion. The pragmatic agents greatly improve the well-formedness of the base agents \emph{ after only one recursion}. As observed for efficiency as well, we see that sRSA, which takes the structure into account, improves the well-formedness to a greater extent. In the limit, see Figure \ref{fig:human_well_inf}, the sRSA agents converge to optimal naming distributions w.r.t. the well-formedness criterion.

Many studies, including the recent one in \citet{F+21}, have reported that humans rarely use more than 1 or 2 levels of recursion in signaling games. It is therefore intriguing that the sRSA only needs only 1 or 2 recursions to reach the information-theoretic frontier. We believe this is something worth exploring further in the future. 


\begin{figure*}
    \centering
    \vspace{-1.5cm}
    \begin{tabular}{cc}
        \subfloat[RSA]{\includegraphics[width=0.49\textwidth]{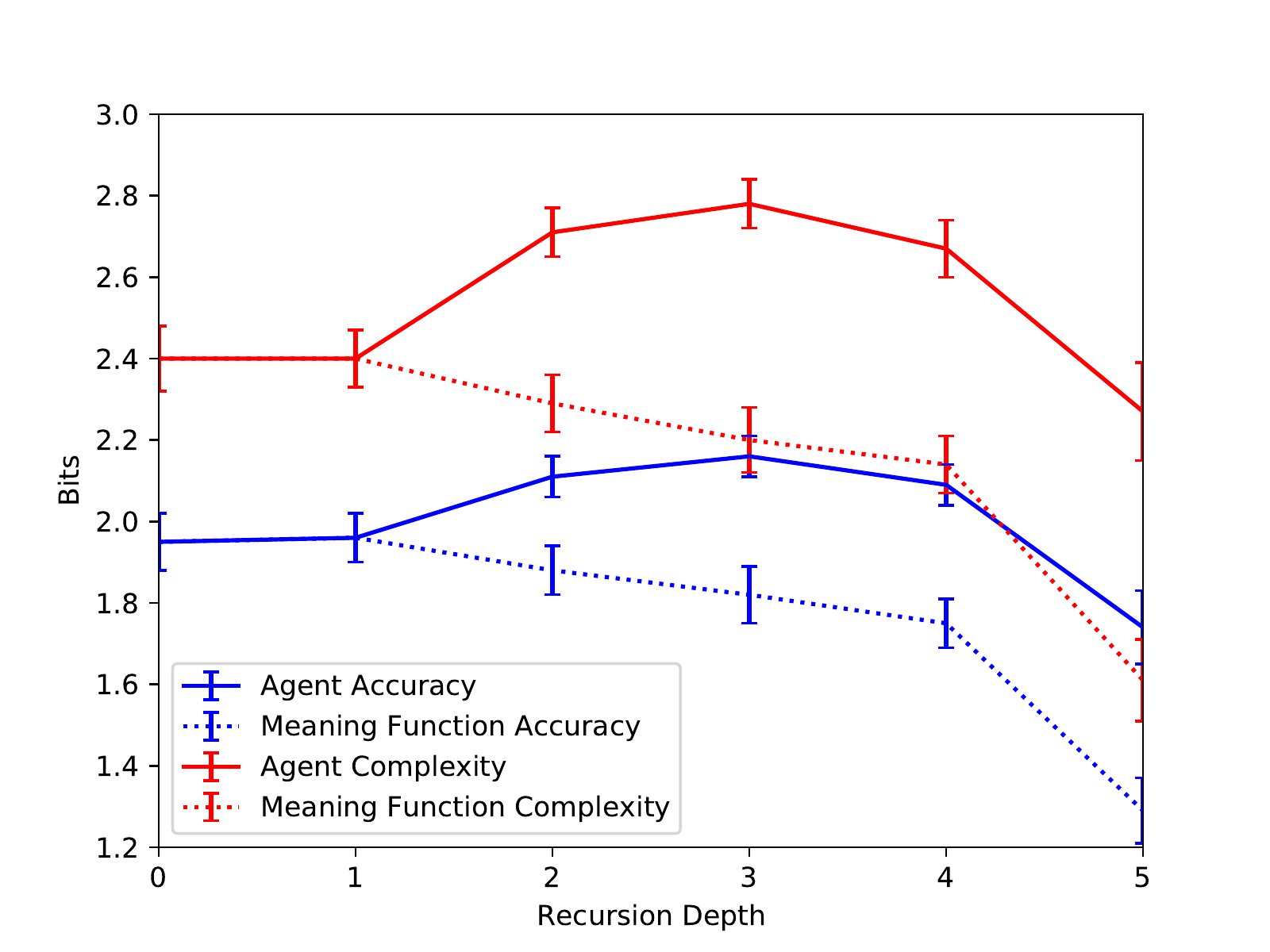}\label{fig:rsa}} & \hspace{-1cm}
        \subfloat[sRSA]{\includegraphics[width=0.49\textwidth]{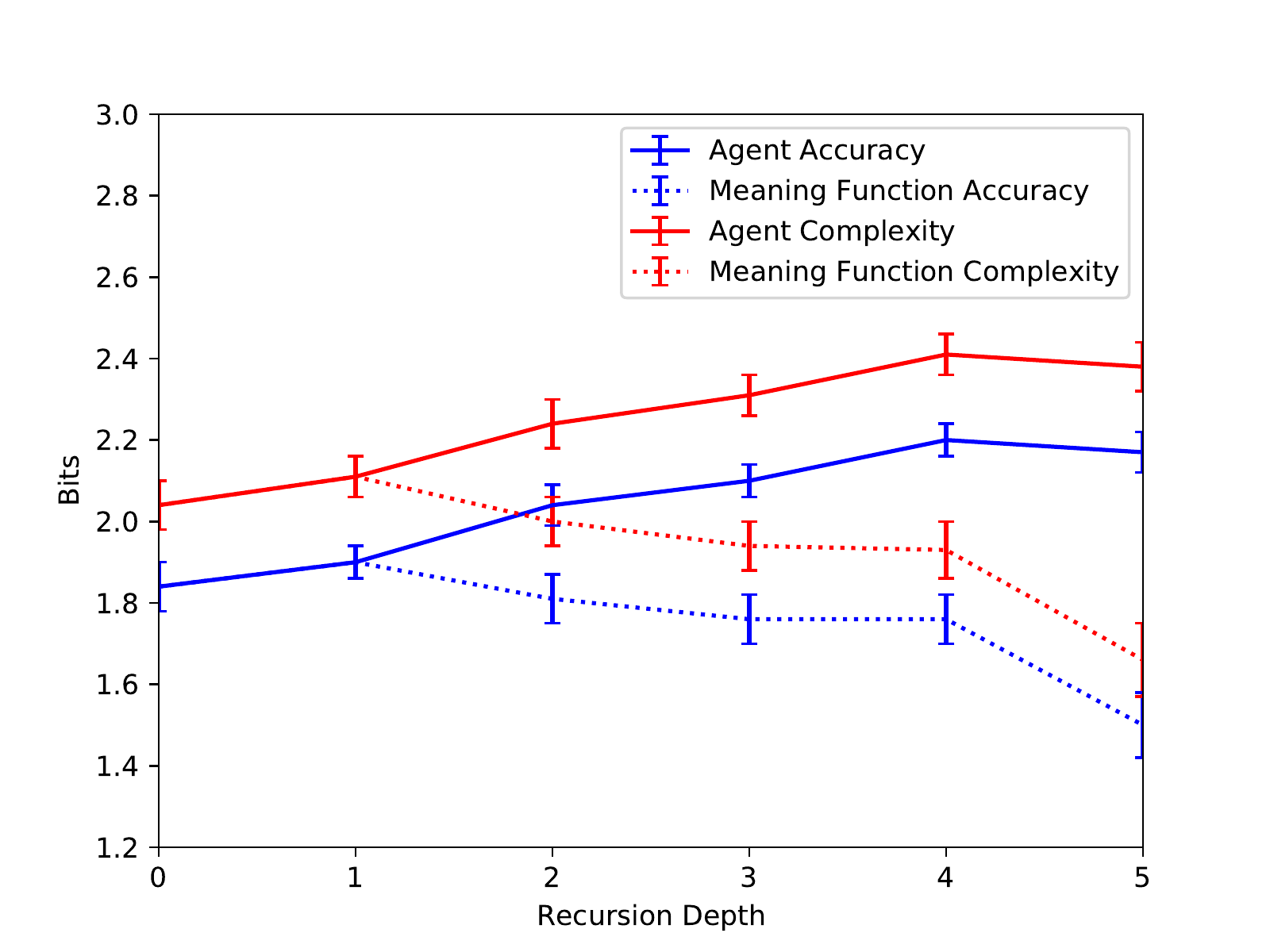}\label{fig:srsa}}
    \end{tabular}
    \caption{In the following plots, depth indicates the level of the final listener in the recursion, and the error bars correspond to the width of the $95\%$ confidence interval. We observe that, as the depth of recursion increases, the accuracy and complexity of the agent differs more compared to the accuracy and complexity of the corresponding meaning function. Noteworthy is that the complexity and accuracy of the sRSA agents increase with recursion depth, while the complexity and accuracy of the corresponding meaning functions decrease. Hence, as the reasoning depth increases, the ambiguity of the learned meaning function increases. The efficiency and accuracy of the agents and meaning functions should be the same at depth $0$ and $1$ since both correspond to the sender $S_{1}(w|m)$.}
    \label{fig:rl_agents}
\end{figure*}

An outlier, when it comes to both efficiency and well-formedness, is the base agent of the language Karajá, highlighted by the black square in Figures \ref{fig:human_IB} and \ref{fig:well_depth1}.  In Figure \ref{fig:Karaja_plot} we illustrate the efficiency and well-formedness of the corresponding RSA and sRSA agents as we increase the recursion depth. Interestingly, applying a few steps of sRSA, see Figure \ref{fig:Karaja_plot}, yields a near-optimal agent, both when it comes to well-formedness and efficiency. This suggests that even though the naming distribution of Karajá is not efficient and well-formed in itself, it serves as a good initialization for a pragmatic and rational agent - but for an agent that takes domain structure into account. Without taking the structure into account, the RSA agent doesn't lead to a more efficient behavior; instead the RSA agent seems to be moving away from the optimal frontier. 

In Figure \ref{fig:Karaja}, we see the corresponding mode-maps for the different RSA versions at depth $1$ and in the limit. We clearly see that taking the structure into account in the reasoning process produces agents that have very smooth mode-maps already at depth $1$, see Figure \ref{fig:Karaja_rkk1}. Here we also see that the standard RSA objective, see Figures \ref{fig:Karaja_rsa1} and \ref{fig:Karaja_rsa101}, fails to produce smooth mode-maps since it does not account for the structure of the domain space. Worth highlighting is that the sRSA, Figure \ref{fig:Karaja_rkk101}, seems to converge to a mode-map very close to the optimal mode-map w.r.t. the well-formedness measure, see Figure \ref{fig:Karaja_ccc}. This is perhaps expected since the sRSA utility considers perceptual similarity. 

\subsection{Artificial Agents}
In our multi-agent reinforcement learning framework, two agents will play a structured signaling game about colors. In the beginning of each game, one agent is randomly assigned to be the speaker agent and the other one acts as a listener. Each agent will keep their own parameterization of the meaning function $\mfunc_\theta$ using a neural network with parameters $\theta$ and $\phi$. Given a context, both agents will apply either RSA or sRSA on the meaning function for $t$ iterations to get their corresponding policies $S_{t, \theta}(w |m, \mathcal{C})$ and $L_{t, \phi}(m|w, \mathcal{C})$. The speaker agent then samples an utterance given the target according to $S_{t, \theta}(w |m, \mathcal{C})$, and upon receiving the utterance, the listener samples a guess according to the distribution $L_{t, \phi}(m|w, \mathcal{C})$. A binary reward is given to both agents depending on whether the listener produced a correct guess and both agents will update their respective meaning function using the REINFORCE objective \citep{Williams}, which for the sender agent corresponds to taking the gradient of $r \log S_{t, \theta}(w |m, \mathcal{C})$
and for the listener gradient of $r \log L_\theta(m|w, \mathcal{C})$. A similar computational setup was recently considered in \citet{Ohmer22}.

We take each neural network to have one hidden layer of $25$ neurons with ReLU activation for the hidden layer and sigmoid activation in the output layer. We train the agents on contexts consisting of $5$ colors sampled from the Munsell chart and represented as a vector in CIELAB space. We vary the depth of the agent from $0$ to $5$, where depth $0$ indicates a sender interacting with a literal listener, and we set $\alpha = 5$. During the evaluation, the context given to the agents will be the entire Munsell chart, as was done for human representations. Each configuration of agents is averaged over $100$ different random seeds. We update the neural networks using standard stochastic gradient descent,  with the learning rate set to $0.001$. The agents were trained for $10 \, 000$ updates using a batch size of $100$.  We compare the results to a pure reinforcement learning baseline (RL) with the meaning function of the same size as that of the pragmatic agents, but with linear activation in the output layer. The RL sender performs a softmax operation over words given a color, and the RL listener performs a softmax operation over colors given a word. This color game is similar to the ones considered in \citet{Kageback2020, Chaabouni21} with the difference that the sender observes the context in our setup.
\begin{figure}[h]
    \centering
    \includegraphics[width=0.49\textwidth]{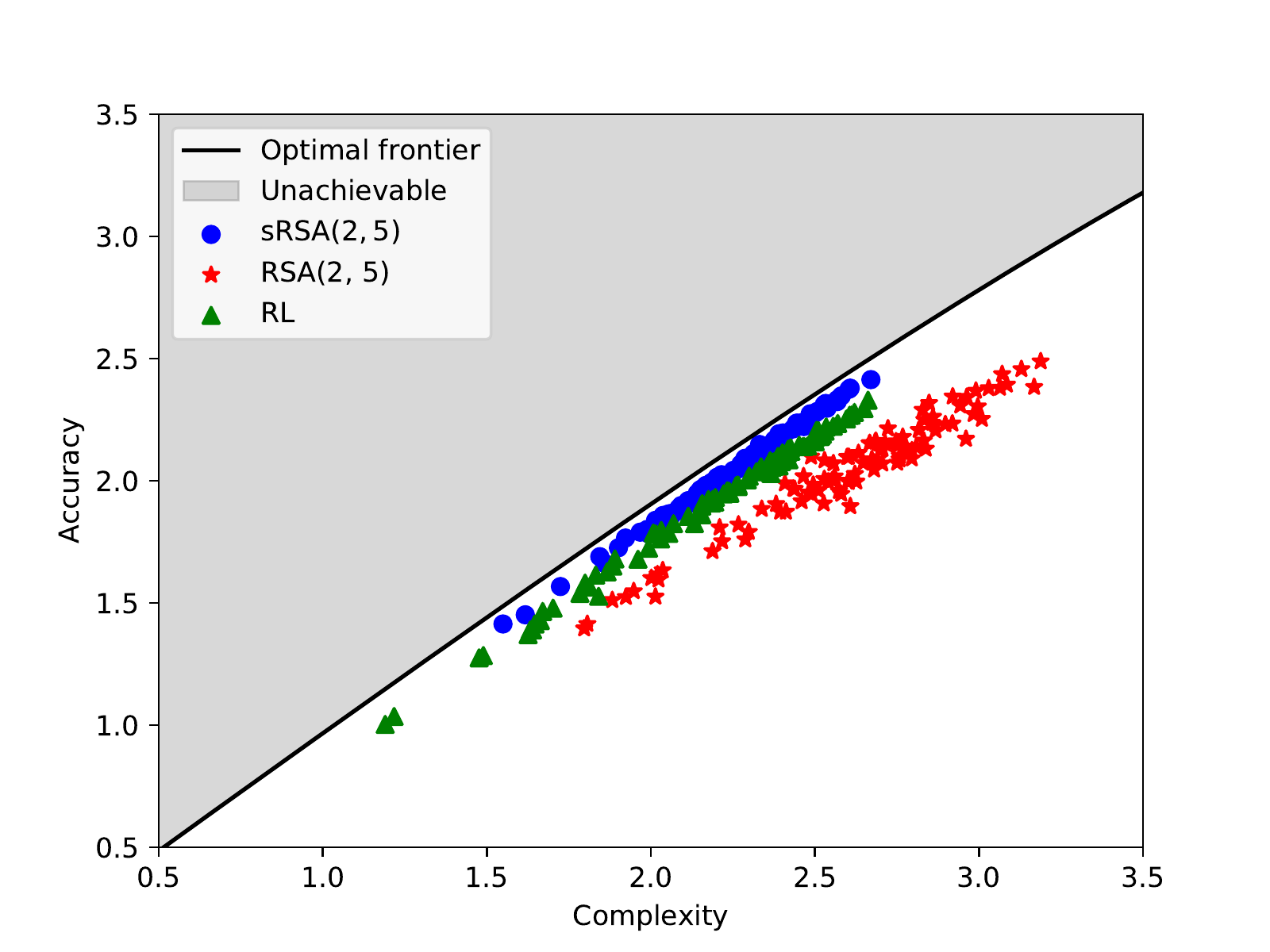}
    \caption{The efficiency of the RSA and sRSA agents trained using a recursion depth of $2$ compared to the RL baseline.}
    \label{fig:agent_eval}
\end{figure}

In Figure \ref{fig:agent_eval}, we observe the efficiency of the agents when performing 2 recursions. The RSA agents develop less efficient communication compared to the sRSA agents and the RL baseline. The sRSA agents develop communication closer to the optimal frontier compared to the RL and RSA agents, illustrating that pragmatic agents with appropriate utility functions develop efficient communication. It is worth highlighting that the RSA and RL agents account for the structure of the color space in their non-contextual meaning functions, i.e. in their neural networks. The results in Figure \ref{fig:agent_eval} thus suggest that the efficiency of the sRSA agents cannot be mimicked by just a graded, or fuzzy, meaning function, but is due to explicitly accounting for the structure in the recursion.
We also note that the non-pragmatic RL baseline learns color naming systems which are more efficient than the pragmatic RSA agents, and that these systems are also close to the information-theoretic frontier (the efficiency of RL agents w.r.t. this objective was first reported in  \citet{Chaabouni21}).

  In Figure~\ref{fig:rl_agents} we see how the complexity and accuracy of the agents and the meaning function changes as the agents are allowed to perform deeper reasoning during learning. As the recursion depth increases, the sRSA agents develop more complex and accurate behavior while ambiguity emerges to a higher extent in the corresponding meaning functions, see Figure \ref{fig:srsa}. Hence, the sRSA agents are able to use ambiguity as a tool to reach greater communicative efficiency. This is consistent with the observations in \citet{Peloquin2020} and the claims in \citet{Piantadosi2012} that ambiguity is associated with efficient communication.  The ambiguity of the meaning function increases with recursion depth, also for the RSA agents, which can be seen in Figure \ref{fig:rsa}. However, for the RSA agents we also observe that the accuracy and complexity of the agent decreases after a few recursions, which seems to indicate that a small number of recursions is better for developing accurate behavior compared to higher recursion depth when using RSA.

\section{Conclusions}
In this work we have explored pragmatic reasoning in a structured signaling game in the color domain. We explored human representations from the World Color Survey, as well as representations learned by artificial agents using reinforcement learning that incorporate pragmatic reasoning. We have seen that, in both cases, incorporating the domain structure in the reasoning process greatly improves the efficiency in the standard information-theoretic sense, compared to using the standard RSA recursion. 

We believe that an interesting future direction is to extend the idea of a structured signaling game and sRSA to more complex environments. An example is a scenario where meanings constitute several different features, and not just one, as considered here. Another interesting future direction, pointed out by one of the reviewers, is to explore scenarios where agents do not share the exact same notion of similarity. 

\section{Acknowledgements}
We thank Terry Regier and the reviewers for providing valuable input on this work. We also want to thank Fredrik D. Johansson, Emilio Jorge and Niklas Åkerblom for providing valuable comments on a previous draft of this paper.

This work was supported by funding from Chalmers AI Research Center (CHAIR) and the computations in this work were enabled by resources provided by the Swedish National Infrastructure for Computing (SNIC).

\bibliography{refs}
\end{document}